\pdfoutput=1

\documentclass[11pt]{article}

\usepackage[final]{acl}

\usepackage{times}
\usepackage{latexsym}

\usepackage[T1]{fontenc}

\usepackage[utf8]{inputenc}

\usepackage{microtype}

\usepackage{inconsolata}

\usepackage{graphicx}

\usepackage{algpseudocode}
\usepackage{algorithmicx}
\usepackage{algorithm}
\usepackage[inline]{enumitem}
\usepackage{graphbox}
\usepackage{microtype}
\usepackage{hyperref}
\usepackage{url}
\usepackage{booktabs}
\usepackage{multirow}
\usepackage{subcaption}
\usepackage{enumerate}
\usepackage{amsmath}
\usepackage{amssymb, bm}

\title{Exploring Quantization for Efficient Pre-Training of Transformer Language Models}

\author{
Kamran Chitsaz$^{1,2}$ \quad Quentin Fournier$^{1,2}$ \quad Gonçalo Mordido$^{1,2,3}$ \quad \textbf{Sarath Chandar}$^{1,2,3,4}$ \\~\\
$^1$Chandar Research Lab \quad
$^2$Mila – Quebec AI Institute \quad $^3$Polytechnique Montréal \quad \\$^4$Canada CIFAR AI Chair \\
\texttt{
\{firstname.lastname\}@mila.quebec}}

\begin{document}
\maketitle
\begin{abstract}
	The increasing scale of Transformer models has led to an increase in their pre-training computational requirements. While quantization has proven to be effective after pre-training and during fine-tuning, applying quantization in Transformers during pre-training has remained largely unexplored at scale for language modeling. This study aims to explore the impact of quantization for efficient pre-training of Transformers, with a focus on linear layer components. By systematically applying straightforward linear quantization to weights, activations, gradients, and optimizer states, we assess its effects on model efficiency, stability, and performance during training. By offering a comprehensive recipe of effective quantization strategies to be applied during the pre-training of Transformers, we promote high training efficiency from scratch while retaining language modeling ability. Code is available at \url{https://github.com/chandar-lab/EfficientLLMs}.
\end{abstract}

\section{Introduction}
\label{sec:intorduction}

Transformers~\citep{vaswani2017attention} have become the dominant model for natural language processing, with the GPT family of models~\citep{radford2019language} showcasing their effectiveness across various tasks, from language understanding to code generation. As the performance of Transformers scales nicely with the number of parameters and the data size, the current state-of-the-art models have reached unprecedented computational requirements both during training and inference~\citep{tay2020efficient}. For instance, pre-training a 175B parameter GPT-3 model requires a staggering number of 10,000 V100 GPUs for 14.8 days~\citep{Patterson2021}. As a result, pre-training has become extremely expensive, beyond the reach of most research groups, and has raised concerns over sustainability due to CO2 emissions from extensive GPU usage~\citep{luccioni2023estimating}.

To improve the efficiency of Transformers, quantization has gained significant traction due to its recent successes in both post-training~\citep{ashkboos2024quarot, dettmers2022gptint, frantar-gptq} and during fine-tuning~\citep{li2023qft}. Quantized pre-training, where certain parts of the computational graph and model parameters are quantized from the beginning of training, remains a challenging problem. In such scenarios, the training instabilities caused by substantial changes in model parameters and emerging model behaviors do not pair well with the added noise introduced by quantization~\citep{pmlr-v162-nagel22a}. Additionally, quantizing model components without compromising performance becomes increasingly difficult at larger scales~\citep{Dettmers2023_4bitcase}. Despite its importance, quantized pre-training of Transformer language models remains largely unexplored at scale.

In this paper, we present the first in-depth study on the effects of quantizing Transformer language models during pre-training and at scale. Our primary aim is to provide a recipe for quantized pre-training by conducting a controlled study that investigates the impact of quantization on weights, activations, gradients, and optimizer states on model efficiency, stability, and performance, using a simple linear quantization with 4 and 8 bits. Our findings demonstrate that 8-bit quantization for weights and activations can be effectively combined to provide significant memory savings and potential speedup, achieving performance comparable to the baseline model. However, extending quantization to gradients to utilize computational speedup in backward matrix multiplications or reducing precision to 4 bits results in notable training instability.

Specifically, 4-bit quantization introduces a sharper loss landscape for weights (\S\ref{sec:weight_quantization}) and persistent outliers in the channel dimension of activations (\S\ref{sec:activation_quantization}), significantly degrading performance despite attempts to manage them through per-channel quantization. Additionally, gradient quantization is particularly problematic due to spikes in gradient norms during early training phases and the unstructured and sparse nature of gradients throughout training (\S\ref{sec:gradient_quantization}). While 8-bit gradient quantization does not hurt model convergence, transitioning to 4 bits results in non-convergence. Additionally, quantizing the first-order moments of Adam to even 4 bits is feasible without significant performance loss, but the second-order moments require a more complex quantization scheme to avoid instabilities in the Adam update, even when using 8-bits quantization (\S\ref{sec:states_quantization}). Lastly, we present our recommended pre-training quantization recipe for the different model components (\S\ref{sec:final_recipe}).

\section{Related Work}
\label{sec:related_work}

In recent years, numerous methods have been studied to improve the efficiency of neural networks. Among these, Quantization-aware training (QAT) emerges as an acceleration technique for inference, as parameters are stored and operations are conducted with higher precision during training. The induced quantization error during training serves as a regularizer, as demonstrated in \citet{gholami2022survey}, ultimately facilitating the development of a more quantization-friendly model.

In contrast, Fully Quantized Training (FQT) harnesses the accelerated gains derived from higher throughput in INT8 or INT4 operations supported by modern GPUs during training. Additionally, FQT capitalizes on memory savings by storing parameters in lower precisions. As exemplified by \citet{li2024memory} and \citet{dettmers20218}, states of the Adam optimizer are stored in 4 and 8 bits, respectively, to minimize memory footprint. Another strategy introduced by \citet{markov2023quantized} involves quantizing both weights and gradients to reduce bandwidth usage in distributed training. Furthermore, \citet{wortsman2024stable} and \citet{kim2021bert} advocate the replacement of linear operations with INT8 matrix multiplications to achieve substantial speedups. Our work extends quantization to multiple components, providing a more comprehensive exploration of the challenges and benefits of quantization during pre-training.

Despite the considerable advantages in accelerating the training process, FQT poses challenges attributable to numerical stability and optimization issues inherent in training quantized networks. Many existing methods predominantly focus on fine-tuning Large Language Models (LLMs) using FQT, leveraging the inherent stability of pre-trained models in each gradient update. The evolution from 16-bit to FP8 data formats, as evidenced by remarkable results in mixed precision training on LLMs~\citep{peng2023fp8}, showcases the potential of FQT. However, the scarcity and difficulty in obtaining hardware that supports FP8 formats pose significant challenges to its widespread adoption.

\citet{wortsman2024stable} employ 8-bit quantization for linear operations in both forward and backward passes, achieving remarkable results in pre-training large-scale vision language models. Their approach incorporates row-wise quantization for activations and gradients, mitigating the impact of quantization errors on other parameters. However, vision language models significantly differ from large textual language models. \citet{xi2024training} explored 4-bit quantization in Fully Quantized Training (FQT) using Hadamard transformations to handle activation outliers and proposed bit splitting to quantize gradients in 4-bit precision. However, their work primarily focuses on fine-tuning, and only relatively small models (60M parameters) are pre-trained on a small dataset (WMT 14 En-De), which hinders the generalizability of their findings to larger language models. Our study explores pre-training at a larger scale, utilizing a bigger model and dataset to thoroughly examine the effects of quantization on various model components.

The generalization of these findings to large-scale language models (LLMs) remains a challenge, especially considering that training such models involves unique complexities. Additionally, their reliance on Hadamard transformations imposes restrictions on activation dimensions, limiting applicability to power-of-two dimensions, a constraint not easily met by recent LLM architectures such as Llama.

In contrast to the aforementioned complex quantization methods, such as quantile quantization and learnable quantization parameters, this study focuses on a more straightforward implementation. The objective is to investigate the effects of quantization and explore the feasibility of training LLMs with full integer operations. The primary goal is to offer detailed insights into quantizing different components of the model. While more sophisticated quantization methods may enhance performance, our work serves as a foundational investigation, providing insights and paving the way for further exploration in quantization for LLMs.

\section{Quantization Methodology}
\label{sec:method}

In this work, we explore various quantization schemes (\S\ref{sec:quantization_method}) and granularities (\S\ref{sec:quantization_granularity}) on model components. This controlled approach during pre-training allows us to examine the impacts on language modeling and downstream task performance.

\subsection{Quantization Scheme}
\label{sec:quantization_method}

We start by introducing the quantization procedure used in our study, which is applied to all linear layers of Transformers. We perform fake quantization, where all values and computations are stored with higher precision, and every quantization operation is followed by de-quantization to introduce quantization error. Simulating low-precision pre-training fits the purpose of this study since we aim to analyze the effects of quantizing different model components without focusing on actual training speedups that can be obtained by implementing custom GPU kernels. 

\begin{figure}[htb!]
	\centering
	\includegraphics[align=b,width=0.99\columnwidth]{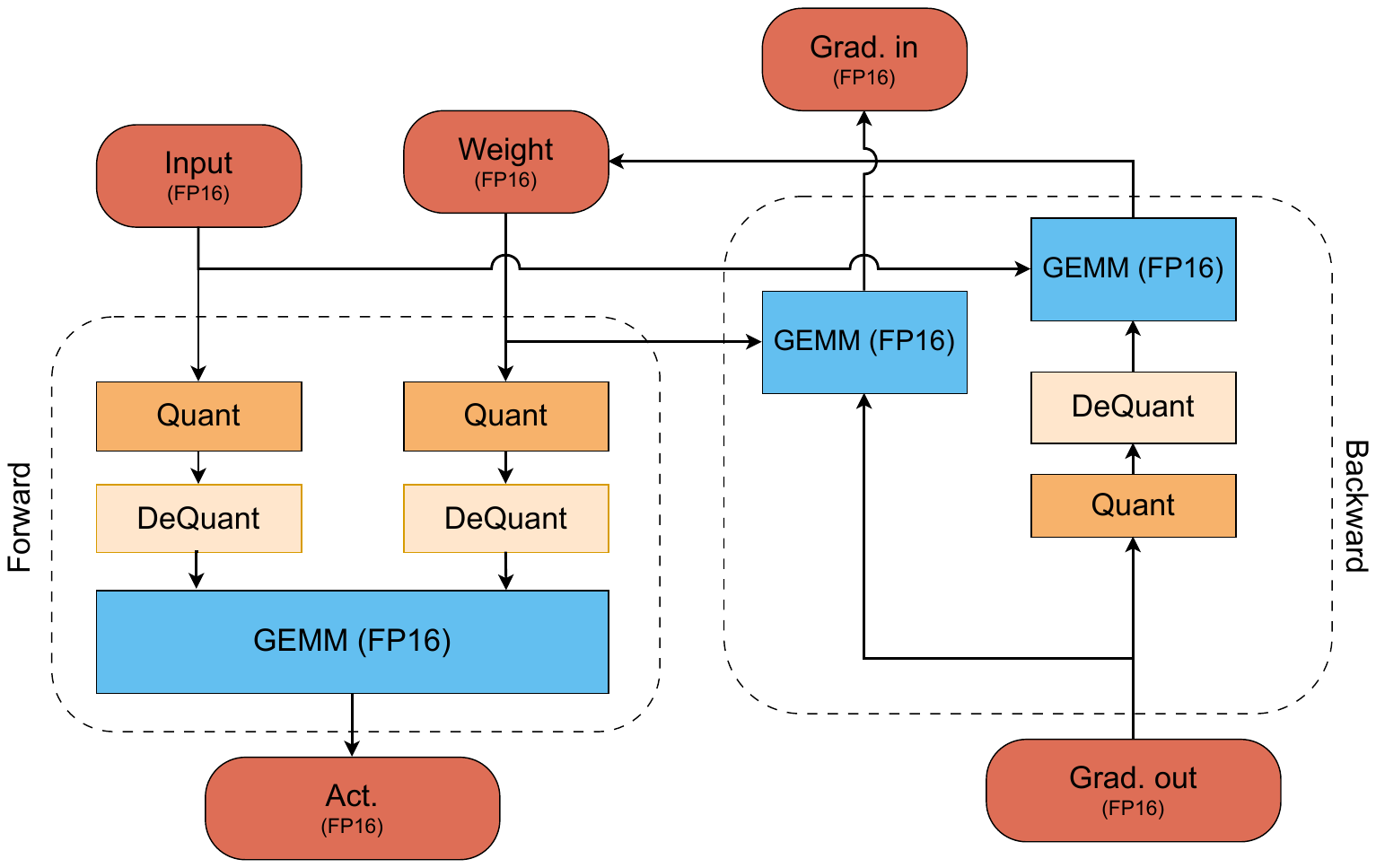}
	\caption{Overview of the quantization process in forward and backward passes.}
	\label{fig:quantization_setup}
\end{figure}

Figure~\ref{fig:quantization_setup} illustrates how quantization error is introduced during both the forward and backward passes. Specifically, the quantization error is injected into either the weights or input activations during the forward pass. Since we only store the gradients related to the weights in memory, the quantization in the backward pass is only applied to the output gradients (``grad out'' in Figure~\ref{fig:quantization_setup}) for weight updates. Nonetheless, the real-valued output gradient is used to compute the quantized input gradient since we observed an increase in training instability when propagating the quantization error through the entire backward path. Additional details about gradient quantization are presented in Section~\ref{sec:gradient_quantization}.

We employ linear quantization for our experiments since this is a popular approach compatible with existing hardware, and using more complex methods could potentially hinder the practical relevance of our study. Specifically, we map real-valued vectors $\boldsymbol{X}$ to a discrete grid of integers as follows:
\begin{equation}\label{eq:quant}
	\begin{split}
		X_\mathrm{int} &= \operatorname{clip}\left(\left\lfloor\frac{X}{s}\right\rceil - z; N, P\right), \\
		\widehat{\bm{X}} &= s (X_\mathrm{int} + z),
	\end{split}
\end{equation}
where $\lfloor\cdot\rceil$ is the round-to-nearest integer operator, $N$ and $P$ represent the quantization range, with $N=-2^{b-1}$, $P=2^{b-1}-1$, and $b$ is the bit width, since we deal with signed data in our experiments. The scaling factor $s$ is set to the maximum absolute value of $\boldsymbol{X}/P$. Unless specified, we perform symmetric quantization by setting the offset $z$ to $0$, which has less overhead than asymmetric quantization where $z$ is set to $\left\lfloor\operatorname{min}(X)/s\right\rceil$ ~\citep{nagel2021white}. During backpropagation, we employ the well-known straight-through estimator (STE)~\citep{bengio2013estimating} mechanism to update the weights.

\subsection{Quantization Granularity}
\label{sec:quantization_granularity}

We can choose scaling factors with different granularity: per-tensor, per-channel, and per-token quantization, where each quantization granularity leads to a specific trade-off between efficiency and performance. Specifically, per-tensor quantization offers the highest efficiency since it performs a single element-wise floating-point multiplication for the de-quantization step. However, since only a single value is used to rescale the entire tensor, performance degradation is likely to occur due to such uniform scaling across the tensor elements.

On the other hand, per-channel and per-token quantization offer a finer-grained scaling, where different scaling factors are tailored to specific tensor element groups (\textit{i.e.} channels and tokens, respectively). Even though such approaches help in terms of performance, they introduce an overhead during the de-quantization step. It is worth noting that, in certain instances, these quantization granularities cannot be efficiently implemented by hardware-accelerated GEMM kernels. For example, using per-channel quantization for both weights and activations can not be efficiently implemented.

\subsection{Quantization Efficiency}
\label{sec:quantization_efficiency}
\begin{figure}[!htb]
	\centering
	\includegraphics[width=0.9\linewidth]{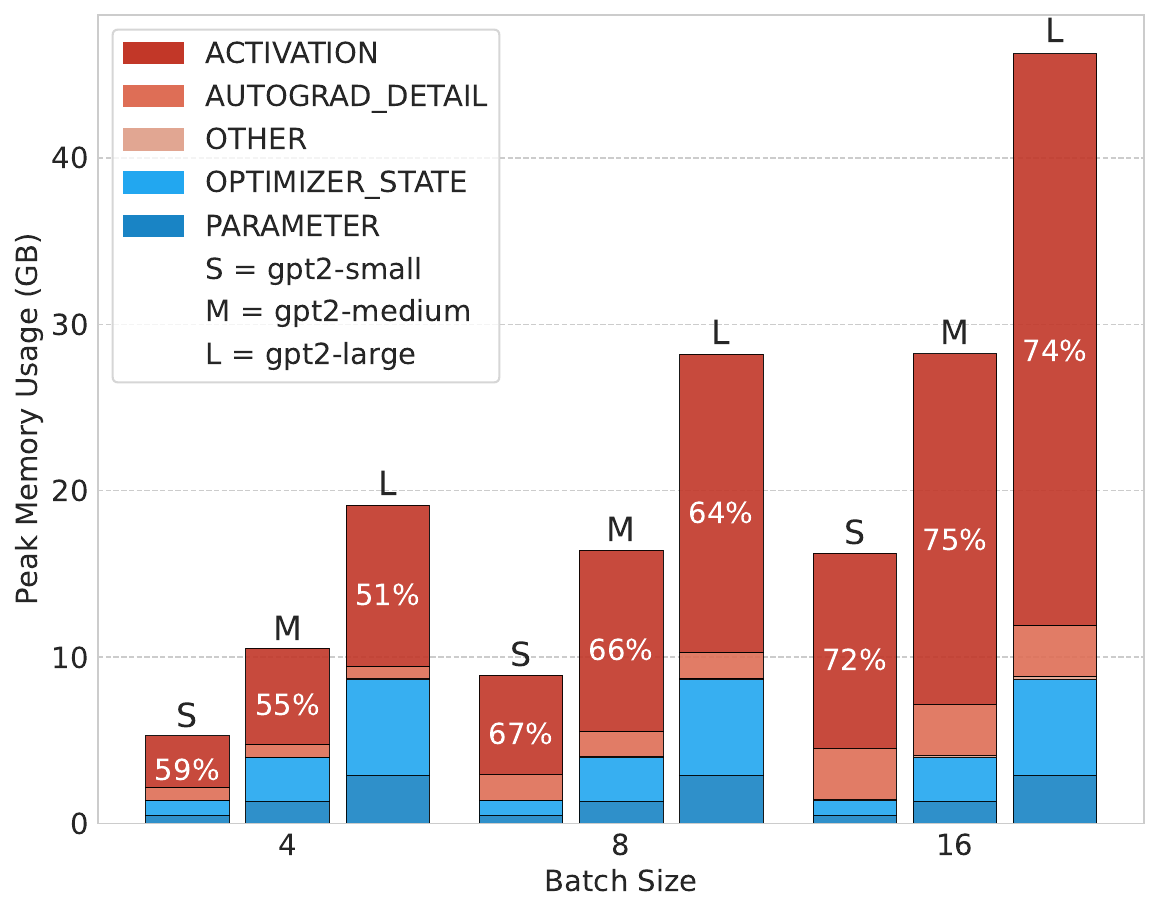}
	\caption{Distribution of peak memory usage across different model sizes (GPT-2 Small, Medium, and Large) for a constant context length of 1024, with varying batch sizes.}
	\label{fig:memory_analysis_compact}
\end{figure}
To show the potential memory saving in quantized pre-training, we explore the memory consumption of various components within GPT-2 models during training using the PyTorch Memory Profiler. We analyze peak memory usage, as shown in Figure~\ref{fig:memory_analysis_compact}, which illustrates memory usage for different batch sizes with a fixed context length of 1024 across various model sizes. We observe that when a model can fit within the GPU memory, the majority of the memory at peak times is consumed by activations, particularly with large batch sizes and sequence lengths. Under these conditions, gradients do not contribute to peak memory usage. More details are in Appendix~\ref{app:memory_analysis}.

\begin{figure}[!htb]
	\centering
	\includegraphics[width=0.9\linewidth]{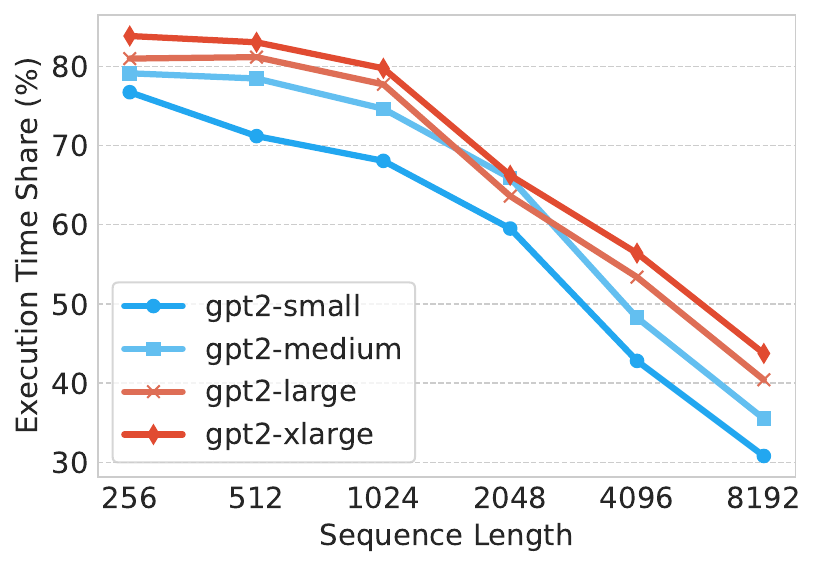}
	\caption{Proportion of total execution time consumed by linear layers in the attention block of GPT-2 models (Small, Medium, Large, and X-Large) across different sequence lengths.}
	\label{fig:time_analysis_compact}
\end{figure}

We also profile the execution time of kernels using the Nvidia Nsight Profiler to assess the potential speedup from quantizing linear layers. Figure \ref{fig:time_analysis_compact} shows the proportion of total execution time consumed by linear layers in the attention block of GPT-2 models of varying sizes across different sequence lengths. This profiling includes both the forward and backward passes. We observe that for small sequence lengths, linear layers consume a significant portion (more than 80\%) of the execution time. As the model size increases, this proportion typically rises, but as sequence lengths increase, the proportion of time spent in linear layers decreases, suggesting that self-attention, due to its quadratic computational complexity, becomes the dominant factor in execution time. This indicates that while quantizing linear layers can offer substantial speedup, the potential gains are more pronounced with smaller sequence lengths.

\section{Experimental Results}

We used GPT-2 small (124M)~\citep{radford2019language} with FlashAttention-2~\citep{dao2022flashattention} for our experiments due to its popularity as a baseline architecture for contemporary studies of LLMs. While larger models are often used, GPT-2 small provides a manageable framework for in-depth experimentation without requiring the immense computational resources of larger models. For our experiments, we pre-trained 30 models from scratch on OpenWebText~\citep{Gokaslan2019OpenWeb} for 300k gradient steps with a global batch size of 512 samples and a context length of 1024 tokens, processing approximately 157 billion tokens, which is consistent with similar works~\citep{liu2023sophia,dao2022flashattention}. For our evaluation setup, we evaluate the performance of the models on a range of language tasks, including ARC-Easy~\citep{yadav2019quick}, ARC-Challenge~\citep{yadav2019quick}, Hellaswag~\citep{zellers2019hellaswag}, LAMBADA~\citep{paperno-EtAl:2016:P16-1}, and GLUE~\citep{wang2018glue}. Additional details about our training and evaluation setups are provided in Appendix~\ref{app:experimentalsetup}.

\subsection{Weight Quantization}
\label{sec:weight_quantization}

\begin{figure}[!htb]
	\centering
	\includegraphics[align=b, width=\columnwidth]{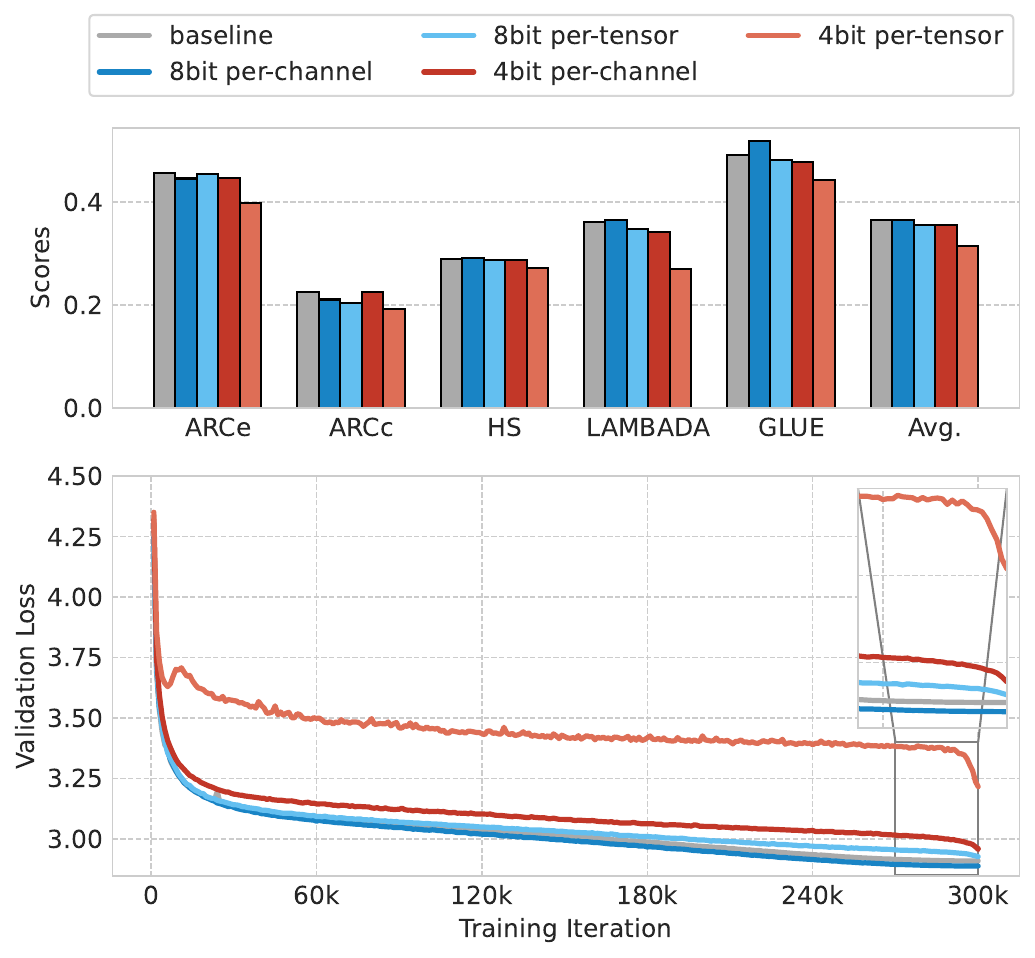}
	\caption{Comparison of different \textbf{Weight Quantization} schemes. (Down) Validation loss across training iterations for 4-bit and 8-bit quantization, both per-tensor and per-channel, alongside the baseline. (Top) PFew-shot accuracy on downstream tasks for the corresponding quantization approaches, demonstrating the efficacy of 8-bit per-channel weight quantization.}
	\label{fig:train_curves_weight}
\end{figure}

The validation loss curves of applying per-tensor and per-channel quantization to weights with 4 and 8 bits are presented in Figure~\ref{fig:train_curves_weight} (down). We observe that per-channel weight quantization with 8 bits outperforms the floating-point baseline since the beginning of training in terms of validation loss, while per-tensor weight quantization with 8 bits shows competitive performance. When quantizing to 4 bits, there is a substantial difference between the different granularities, with per-channel weight quantization significantly outperforming per-tensor quantization, as previously discussed in \S~\ref{sec:quantization_granularity}.

We also evaluate the downstream task performance of the quantized pre-trained models in Figure~\ref{fig:train_curves_weight} (top). We observe that 8-bit weight quantization outperforms 4-bit and achieves competitive performance compared to the floating point baseline, independently of the granularity used. Once again, per-channel weight quantization with 8 bits achieves the best performance among the tested quantization schemes. Overall, we observe similar findings when comparing the performance of the different methods during the pre-training and downstream phases. Despite the success in quantizing weights to 8 bits from scratch, we note that only performing 8-bit quantization post-training also works well, as shown in Appendix~\ref{app:quantization_results} (Table~\ref{tab:evaluation-ckps-ptq}). However, when it comes to 4-bit quantization, applying quantization from scratch leads to significantly better performance.

\begin{figure}[!htb]
	\centering
	\includegraphics[align=b,width=0.7\columnwidth]{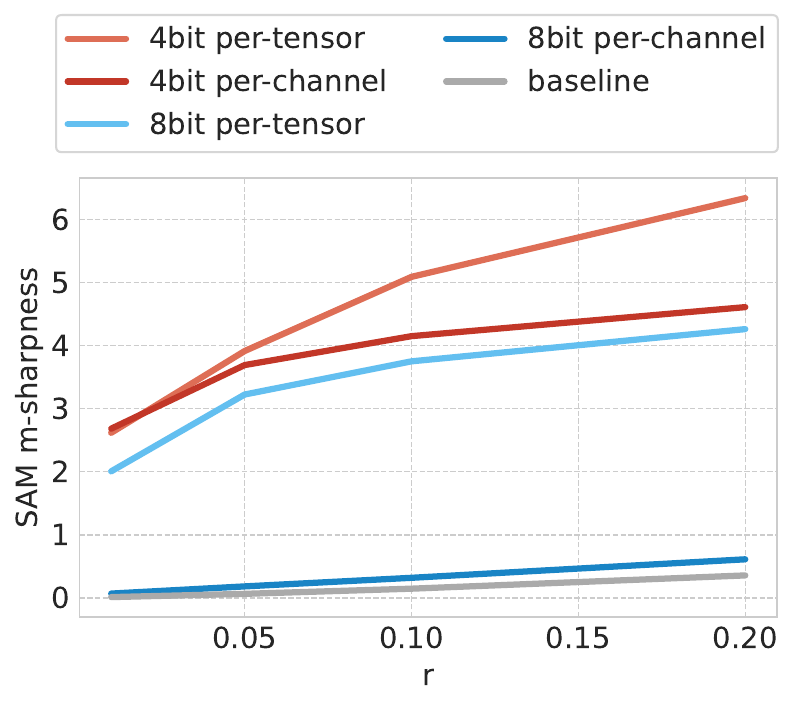}
	\vfill
	\includegraphics[align=b,width=0.9\columnwidth]{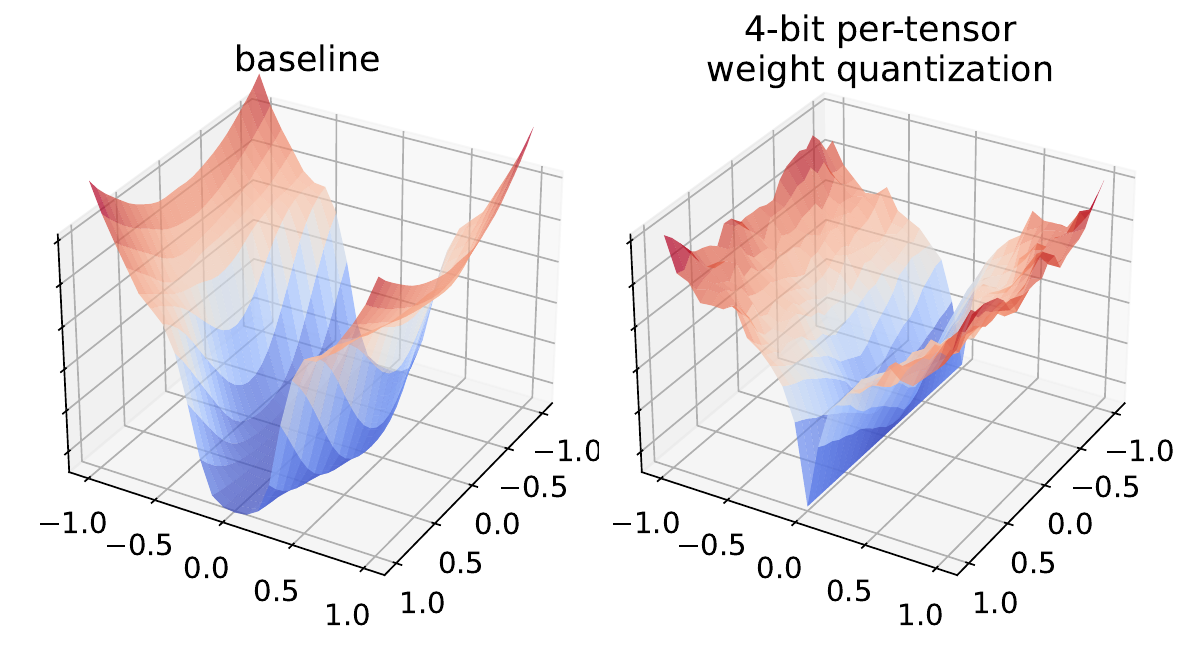}
	\caption{Sharpness comparison between baseline model and 4-bit weight quantization. (Top) $m$-sharpness. (Down) Loss surfaces.}
	\label{fig:weight_sharpness}
\end{figure}

We note a pronounced drop in validation loss at the end of training with the 4-bit weight quantization schemes, as seen in the final training iterations in Figure~\ref{fig:train_curves_weight} (down). We hypothesize that this is related to reducing the learning rate below $1e-6$ in the final steps of training in our setup. However, this is not observed in all the schemes. \citet{pmlr-v162-nagel22a} suggested that the oscillations in weights originated from performing quantization with STE during training may result in weight movements around decision thresholds. Hence, in our use case, we hypothesize that the drop in validation loss stems from the presence of sharp minima when quantizing weights to lower bit-widths and that the lower learning rate regime at the end of training helps convergence to minima with lower loss.

To further investigate this, we compare the sharpness of the different models at the end of pre-training using $m$-sharpness~\citep{foret2021sharpnessaware} with varying radii in Figure~\ref{fig:weight_sharpness} (down). We observe that all quantized models converge to sharper minima compared to the floating-point, unquantized baseline. Moreover, there is a direct relation between the sharpness of each quantized model and the relative drop in validation loss observed in Figure~\ref{fig:train_curves_weight} (down). Specifically, per-tensor weight quantization with 4 bits shows the highest sharpness and also the highest drop in validation loss. This correlation is also observed on a smaller scale for the per-tensor weight quantization model to 4 bits and the per-tensor weight quantization model to 8 bits. To visualize the loss surfaces, we employ the visualization method introduced by \citet{li2018visualizing}. The loss surfaces of the baseline and the per-tensor weight quantization model to 4-bits are shown in Figure~\ref{fig:weight_sharpness} (right), further illustrating the impact that quantization during pre-training has on the sharpness of the final pre-trained model.

\subsection{Activation Quantization}
\label{sec:activation_quantization}

\begin{figure*}[!htb]
	\centering
	\includegraphics[width=0.96\linewidth]{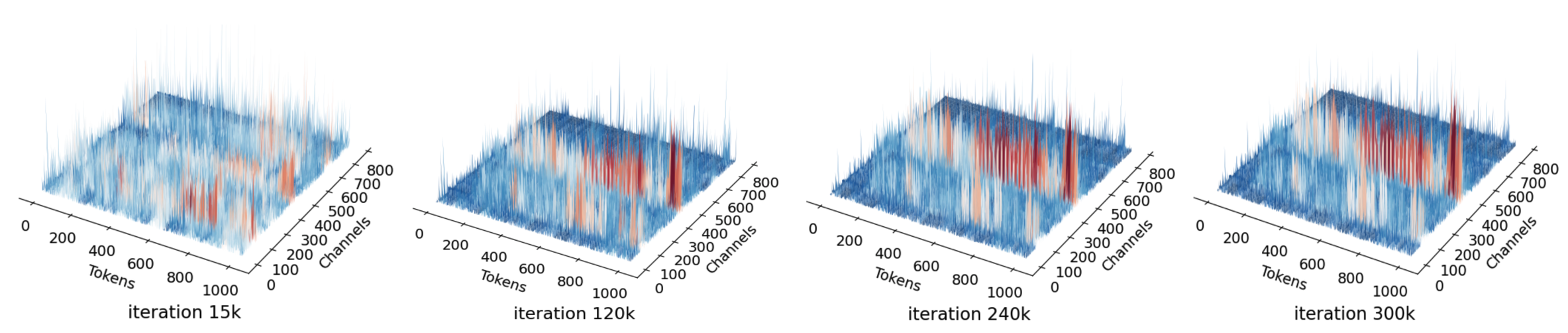}
	\caption{Training progression of activation distributions across selected iterations, showing persistent channel-specific outliers}
	\label{fig:activation_histogram_training}
\end{figure*}

\begin{figure}[!htb]
	\centering
	\includegraphics[width=\columnwidth]{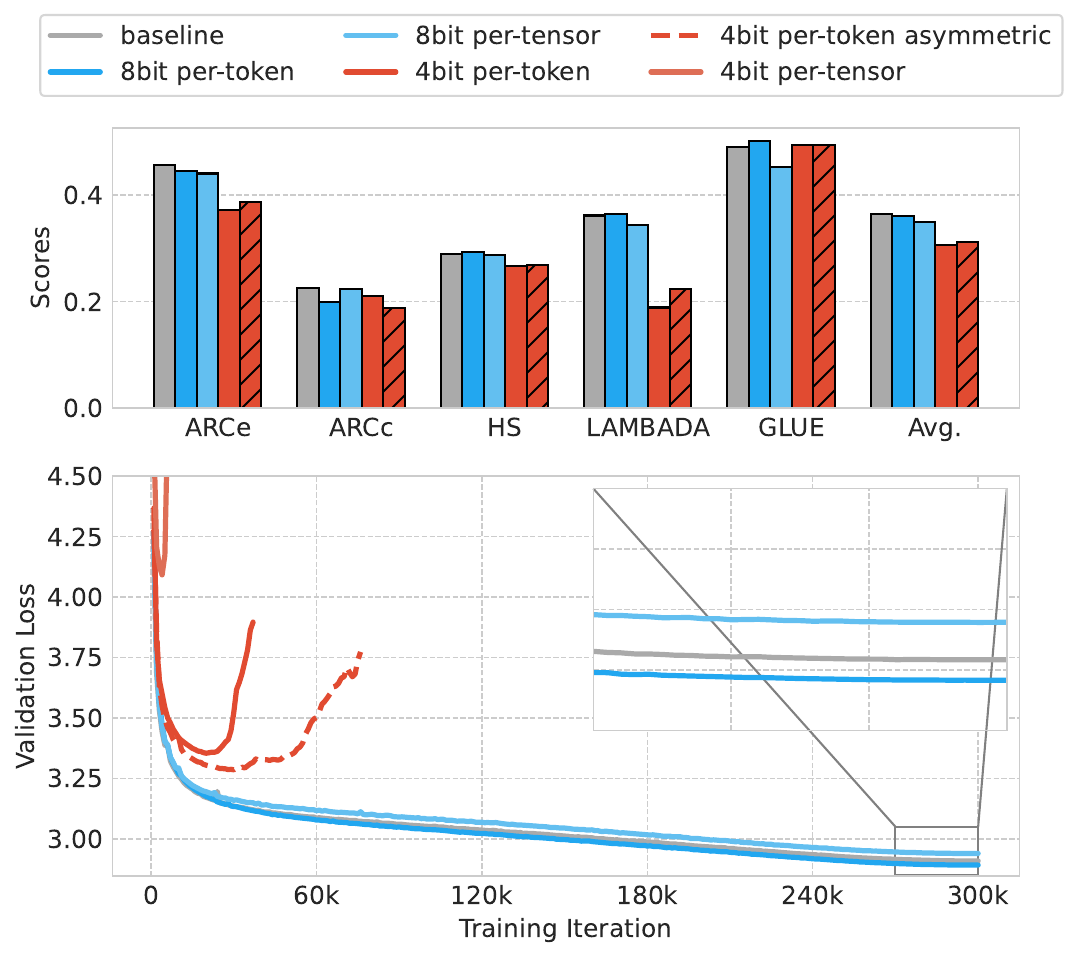}
	\caption{Pre-training \textbf{Activation Quantization} effects: (Down) Validation loss curves for various quantization schemes; (Top) Few-shot accuracy on downstream tasks, showing 8-bit quantization closely aligns with baseline performance.}
	\label{fig:train_curves_activation}
\end{figure}

We present the validation loss when quantizing activations during pre-training in Figure~\ref{fig:train_curves_activation} (down). We note that quantizing activations to 4 bits is more challenging than training with 4-bit weights, as noticed by previous work~\citep{xiao2023smoothquant} and as seen by the divergence behavior of per-token and per-tensor activation quantization. On the other hand, quantizing activations to 8 bits works well, especially if performed per-token, which achieves lower validation loss than the floating-point baseline. The performance of downstream tasks of the baseline and quantized models is presented in Figure~\ref{fig:train_curves_activation} (top). We notice a similar performance trend across the models, with 8-bit per-token activation reaching competitive performance with the baseline model. 

To investigate 4-bit activation quantization further, we also tried applying an asymmetric scheme. The intuition is that, while most activations exhibit symmetry around zero, this is not the case for activation after GELU activation functions~\citep{hendrycks2016gaussian}. Hence, having an asymmetric scheme can lead to a better utilization of the available bits for representation. However, we observe that while the asymmetric scheme provides an improvement over 4-bit per-token symmetric quantization, the model still diverges. We analyze the activation distributions of the output projection layer within the attention block of layer 7 in Figure~\ref{fig:activation_histogram_training}. We see that outliers predominantly reside within specific channels and persistently affect the same channels throughout training. Given our use of per-tensor and per-token quantization for activations, it is evident that such outliers can influence all tokens, given their consistent pattern across the channel dimension.

\begin{figure}[!htb]
	\includegraphics[align=b,width=0.5\columnwidth]{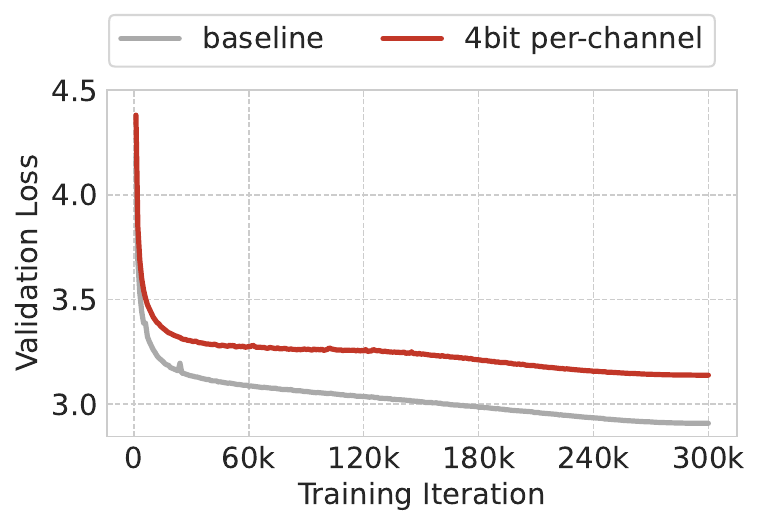}
	\hfill
	\includegraphics[align=b,width=0.4\columnwidth]{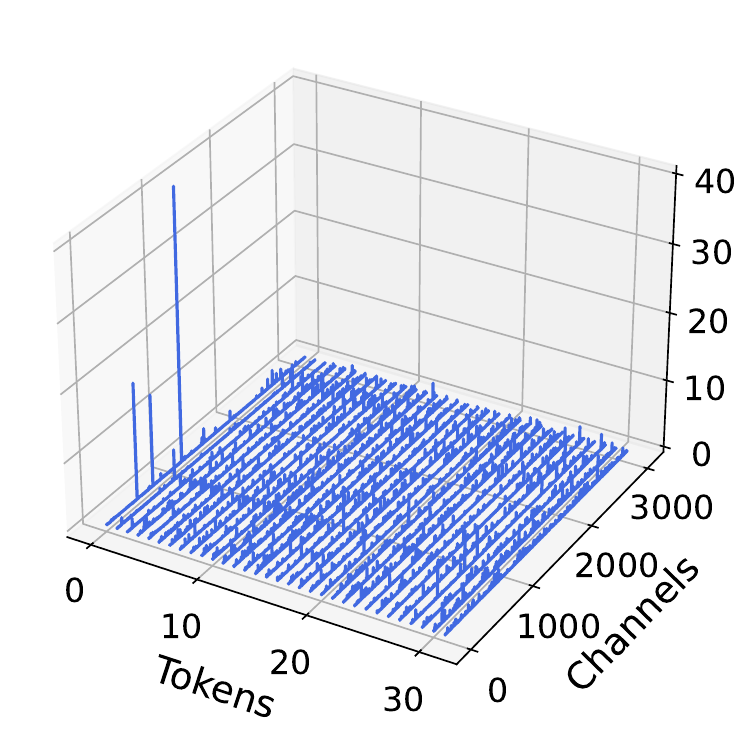}
	\caption{Validation loss and activation outliers for 4-bit per-channel quantization. (Left) Validation loss shows convergence but underperforms compared to the baseline. (Right) Histogram of activation with massive outliers.}
	\label{fig:masive_outliers}
\end{figure}

Since activation outliers are mostly predominant in particular channels during training, we explore the efficacy of per-channel 4-bit activation quantization in Figure~\ref{fig:masive_outliers} (left). We observe that this variant does converge, even though it fails to be competitive with the floating-point baseline. Such degradation in validation loss can be attributed to the presence of massive outlier activations in specific layers. Despite being important for the model's performance~\cite{sun2024massive}, these big activations pose a challenge for both per-token and per-channel quantization. An example of these is presented in Figure~\ref{fig:masive_outliers} (right), showing the presence of large activations in the FC2 layer in the final attention block.

\subsection{Gradient Quantization}
\label{sec:gradient_quantization}

\begin{figure}[hb!]
	\centering
	\includegraphics[width=\columnwidth]{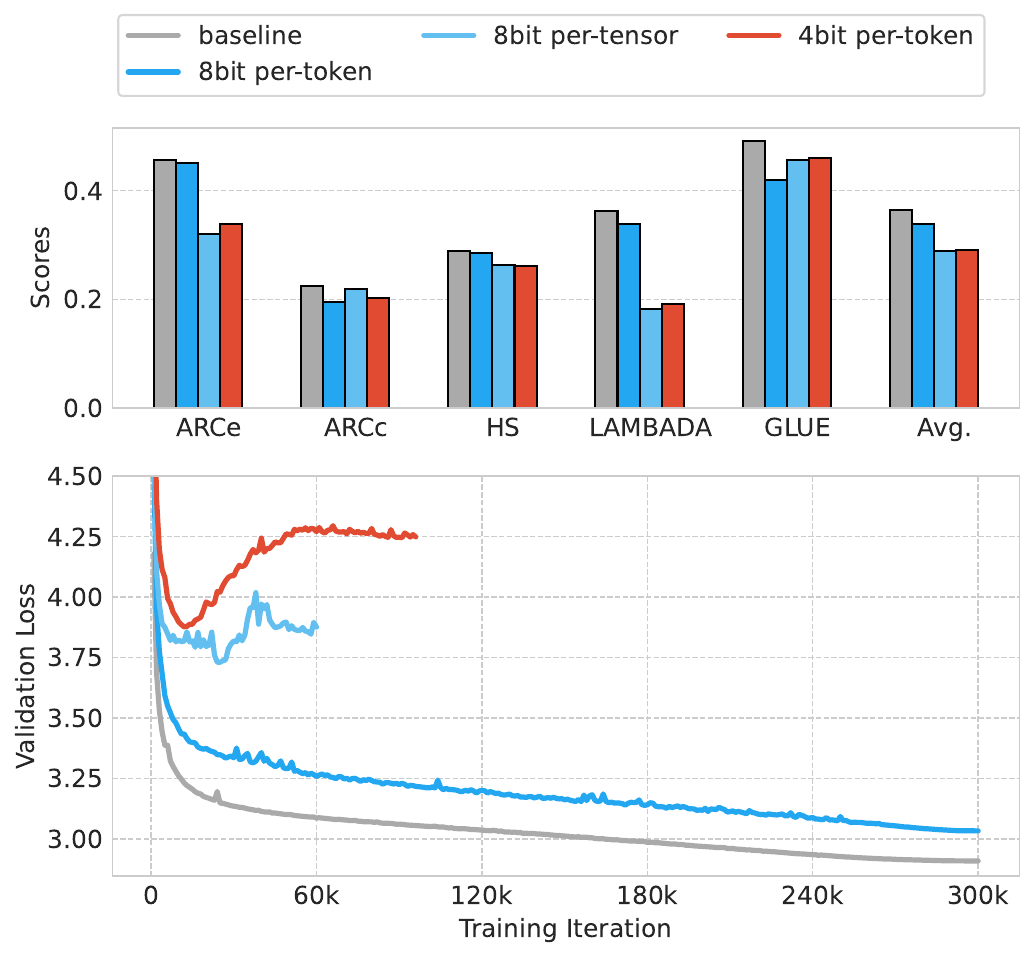}
	\caption{\textbf{Gradient Quantization}. (Down) Validation loss showing non-convergence for 4-bit and 8-bit per-tensor quantization. (Top) Few-shot accuracy on downstream tasks, with only 8-bit per-token approaching baseline performance.}
	\label{fig:train_curves_grad}
\end{figure}

We perform gradient quantization and present the validation losses obtained during pre-training in Figure~\ref{fig:train_curves_grad} (down). We observe that, with 4-bit gradient quantization, training becomes highly unstable or completely fails to converge. With 8 bits, only per-token quantization converges despite showing worse performance compared to our baseline model. The observations are similar when measuring the performance of the different models on downstream tasks, as shown in Figure~\ref{fig:train_curves_grad} (top).

\begin{figure}[!htb]
	\centering
	\includegraphics[width=\columnwidth]{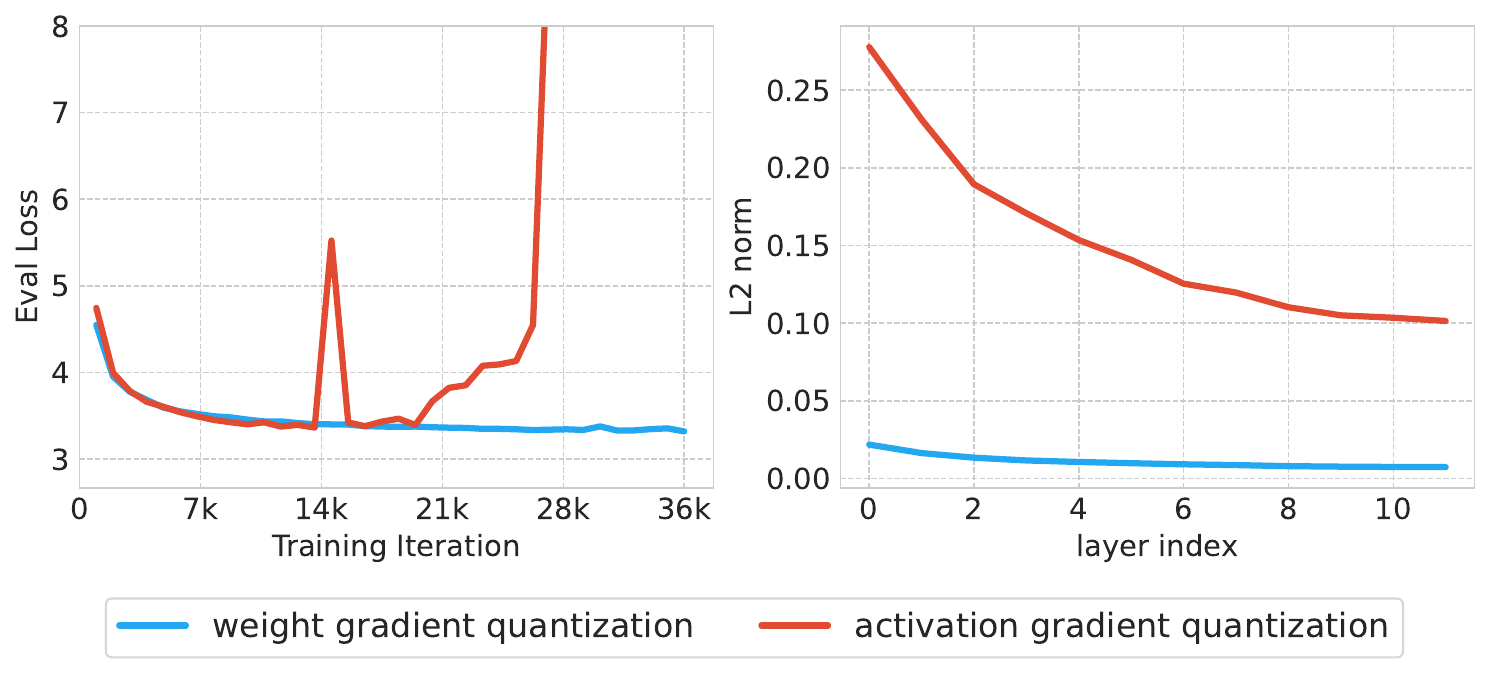}
	\vfill
	\includegraphics[width=0.8\columnwidth]{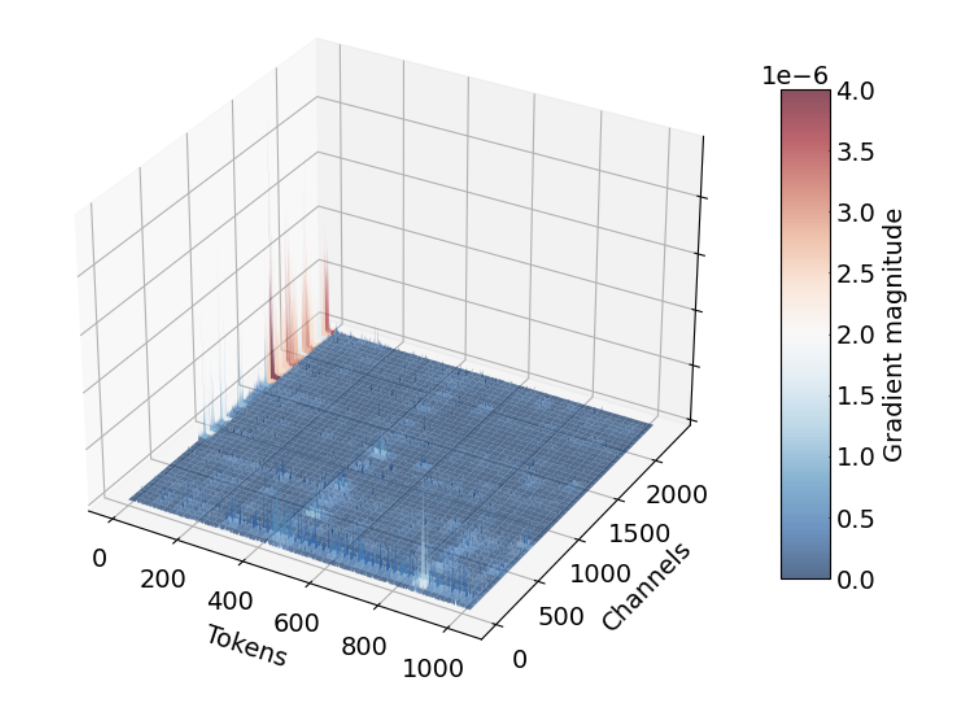}
	\caption{(Top) Validation loss spike illustrating instability from quantizing activation gradients. (Down) Gradient magnitude histogram for a linear layer, highlighting sparsity and potential for quantization error.}
	\label{fig:gradient_analysis}
\end{figure}

As previously discussed in \S\ref{sec:quantization_method}, we quantize the output gradients only for the weight updates, avoiding the instability in training that can result from propagating quantization errors when quantizing activation gradients. This is illustrated in Figure~\ref{fig:gradient_analysis} (top), where quantizing activation gradients in the initial training stages leads to an explosion in the validation loss, followed by divergence. Moreover, we observe an increase in the L2 norm between the floating-point gradients and the quantized counterparts when quantizing activation gradients compared to weight gradients.

To further analyze the subpar performance of 8-bit per-token gradient quantization, even when only applied to weight gradients, we analyze the gradients for the QKV projection at the first layer of the model early on in training in Figure~\ref{fig:gradient_analysis} (down). We observe that gradients are mostly sparse during training and are prone to induce high quantization errors, rendering instabilities.

\subsection{Optimizer States Quantization}
\label{sec:states_quantization}

\begin{figure}[!htb]
	\centering
	\includegraphics[width=\columnwidth]{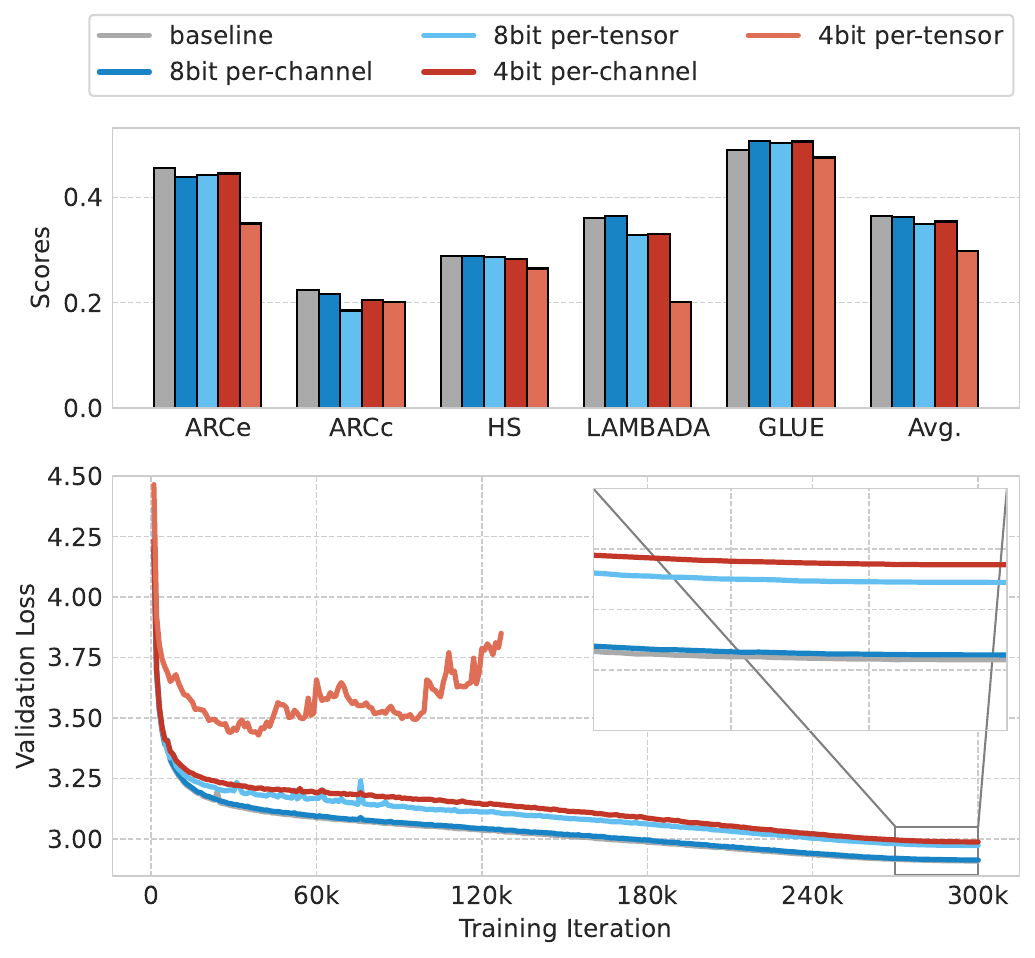}
	\caption{Quantization of \textbf{Adam optimizer}'s first-order moments. (Down) Comparison of validation loss for different quantization approaches; (Top) Few-shot accuracy on downstream tasks, with 8-bit per-channel closely matching baseline.}
	\label{fig:train_curves_first_state}
\end{figure}

We quantize optimizer states, particularly the first and second moments in the Adam optimizer. Specifically, the quantized values of each state are stored until the next training iteration, which are then dequantized and used for Adam's update. To better assess the effect of quantizing each state, we quantize them separately and individually. The validation losses when quantizing Adam's first state are presented in Figure~\ref{fig:train_curves_first_state} (down). We observe that per-channel quantization to 8 bits works well, achieving performance similar to that of the baseline model. Notably, only per-tensor quantization to 4 bits failed to converge out of all tested configurations. Similar findings are found when evaluating the downstream performance of the different quantization schemes in Figure~\ref{fig:train_curves_first_state} (right).

\begin{figure}[!htb]
	\centering
	\includegraphics[width=\columnwidth]{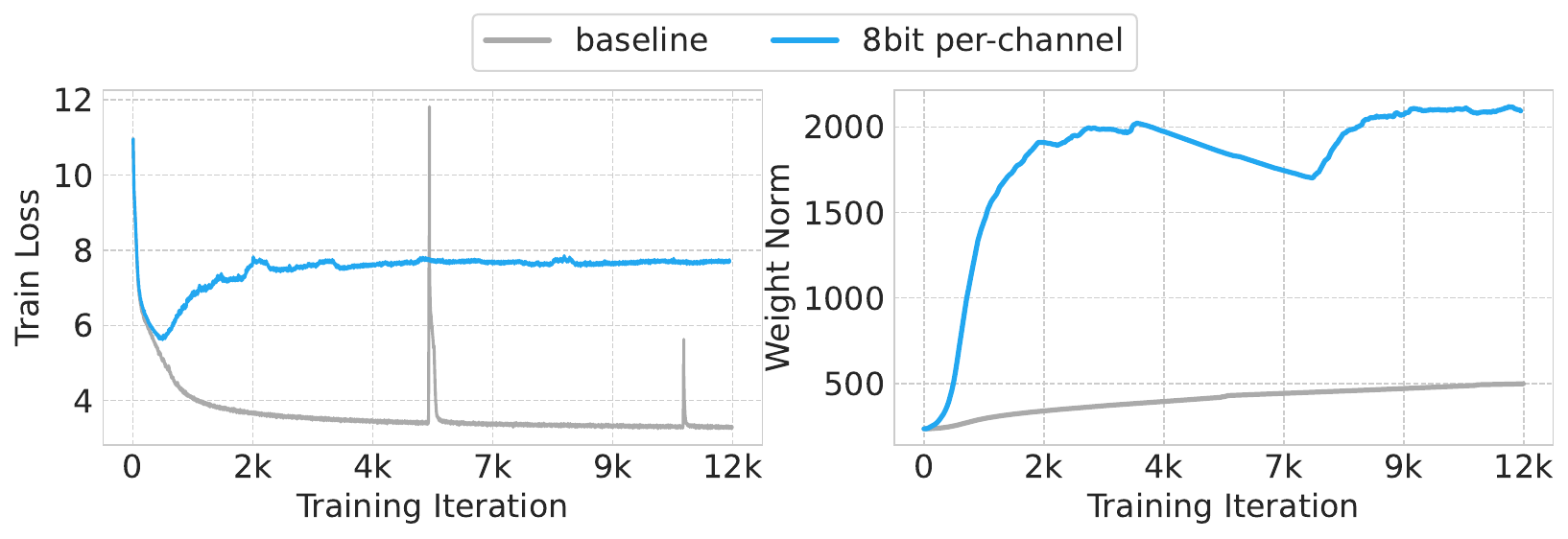}
	\vfill
	\includegraphics[width=0.65\columnwidth]{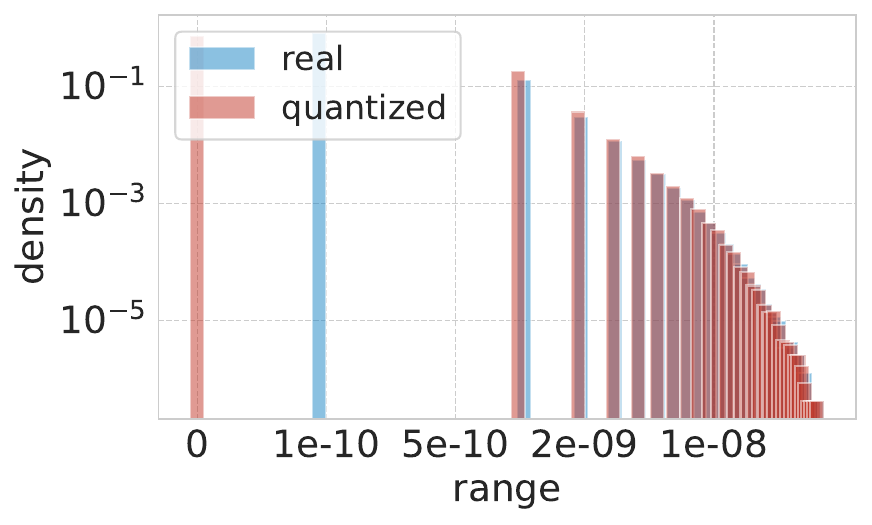}
	\caption{Second-order moments of Adam quantization. (Top) Validation loss quickly diverges. (Down) Histogram showing the concentration of quantized values in the zero bin, which contributes to instability in weight updates.}
	\label{fig:adam_second_state_analysis}
\end{figure}

The results of quantizing Adam's second state are presented in Figure~\ref{fig:adam_second_state_analysis} (left). We observe that the quantized model failed to converge smoothly throughout training, even when applying per-channel quantization with 8 bits. This can be explained by the usage of a linear symmetric quantization function around zero in our scheme. This causes all small values to be set to zero after quantization, hurting performance, as presented in Figure~\ref{fig:adam_second_state_analysis} (right). Given that the second state plays a pivotal role in the denominator of Adam's update, such clustering to zero leads to excessively large weight updates, causing training to diverge from the onset, as observed in Figure~\ref{fig:adam_second_state_analysis}  (left).

\subsection{Multiple Components Quantization}
\label{sec:final_recipe}

\begin{figure}[!htb]
	\centering
	\includegraphics[width=\columnwidth]{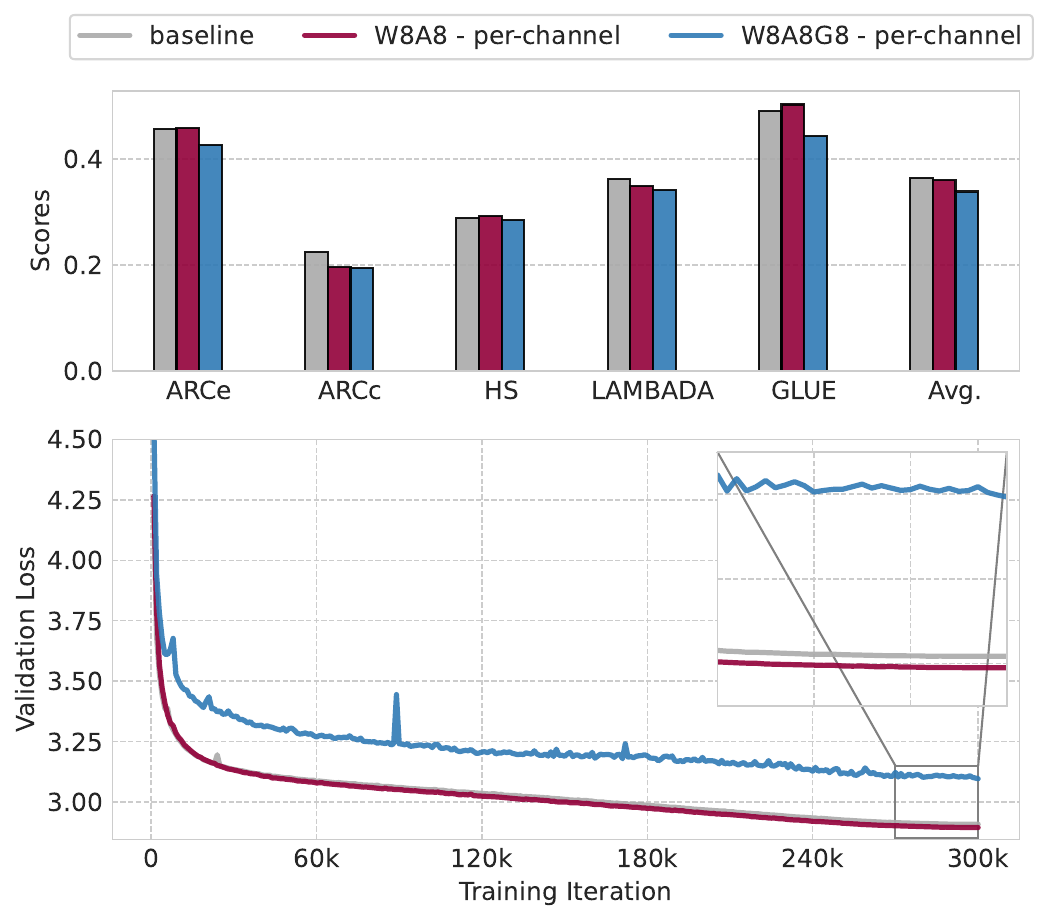}
	\caption{(Down) Validation loss across training iterations for weight, activation, and gradient quantization together; (Top) Few-shot accuracy on downstream tasks.}
	\label{fig:train_curves_all}
\end{figure}

As our last studied components, we trained models using 8-bit quantization for weights, activations, and gradients, as well as isolating quantization to only weights and activations with per-channel granularity for weights and per-token granularity for activations and gradients. Our findings, as depicted in Figure~\ref{fig:train_curves_all}, demonstrate that quantizing both weights and activations to 8-bit allows performance to closely align with the baseline model. However, extending quantization to include gradients results in a notable decrease in performance. This observation is consistent with earlier results from our independent quantization of gradients, highlighting the significant challenges this introduces.

\section{Conclusion}

This study presents an extensive analysis of the impact of quantizing specific Transformer components in 4 and 8 bits during pre-training, in contrast to concurrent work that focuses on individual methods without comprehensive ablations~\cite{xi2024jetfire}. Our study reveals that quantizing weights to 8 bits from the beginning of pre-training is generally successful. However, 4-bit weight quantization can significantly affect model convergence due to a sharper loss landscape. We also found that carefully managing activation outliers is crucial to avoid performance drops with lower-precision quantization. Additionally, we explain the sensitive nature of gradient quantization and its potential to fail at lower bit-widths. We observed that while the first-order moments of the Adam optimizer can be effectively quantized to 4 bits, the second-order moments pose a greater challenge even for 8-bit quantization. Overall, our work establishes an important foundation for future developments in quantization approaches tailored to Transformers, opening the door to efficiently train large-scale models from scratch for improved accessibility.

\section*{Limitation}

We acknowledge the limitations of our work:
\begin{itemize}
	\item We employed linear quantization for our experiments. While this approach is widely used and allows for a controlled study, it may not capture the full potential of more sophisticated quantization methods.
	\item Due to the cost of pre-training and the number of experiments, we limited our study to GPT-2 small, and our findings may not generalize to larger models.
	\item The efficiency gains discussed are estimated using profiling data, and implementing these improvements in practice is challenging due to the complexity of kernel optimizations.
\end{itemize}

\section*{Acknowledgements}

Sarath Chandar is supported by the Canada CIFAR AI Chairs program, the Canada Research Chair in Lifelong Machine Learning, and the NSERC Discovery Grant. Gonçalo Mordido was supported by an FRQNT postdoctoral scholarship (PBEEE) during part of this work. The authors acknowledge the computational resources provided by the Digital Research Alliance of Canada.

\bibliography{custom}

\appendix

\section{Experimental Setup}
\label{app:experimentalsetup}

\subsection{Model and Training Configuration}

Our experiments leverage the computational enhancements of the FlashAttention library~\citep{dao2022flashattention}, utilizing its GPT-2 implementation within the HuggingFace Trainer framework \footnote{\url{https://huggingface.co/docs/transformers/en/main_classes/trainer}} for our training processes. 

Training was conducted on the OpenWebText corpus~\citep{Gokaslan2019OpenWeb}, adopting a set of training configurations similar to~\citep{dao2022flashattention} and nanoGPT~\footnote{\url{https://github.com/karpathy/nanoGPT}}, without hyperparameter tuning due to computational constraints. We use AdamW optimizer with a learning rate of $6e-4$, combined with a cosine learning rate scheduler set to a half cycle. We adopt mixed precision training in bfloat16. The experiments are conducted with a fixed batch size of 512, employing gradient accumulation as necessary to accommodate the computational constraints of our setup across 4xA100 80G GPUs. This training configuration remains consistent across all experiments, culminating in a training duration averaging 4.3 days for completion of 300k steps.

\subsection{Data and Evaluation Metrics}
The OpenWebText corpus was randomly split into training and validation sets, with 0.5\% of the data reserved for validation. Following the methodology of \citet{radford2019language}, the performance of our pre-trained models was evaluated on several benchmark datasets. These evaluations focused on perplexity measurements on well-known datasets including WikiText~\citep{merity2016pointer}, PTB~\citep{marcus-etal-1993-building}, and 1BW~\citep{chelba2014billion}. Our models were further evaluated using a range of downstream tasks:

\begin{itemize}
	\item GLUE~\citep{wang2018glue}: A collection of natural language understanding tasks including question answering, sentiment analysis, and textual entailment, designed to benchmark the generalization capabilities of models across a diverse range of linguistic challenges.
	\item ARC~\citep{yadav2019quick}: Comprising the ARC-Easy and ARC-Challenge, these datasets test the model's reasoning ability through science-based question answering. ARC-Easy contains simpler questions, while ARC-Challenge includes more complex queries demanding deeper reasoning.
	\item Hellaswag~\citep{zellers2019hellaswag}: This dataset challenges models to predict the most plausible continuation of a narrative from a large corpus of everyday contexts and movie scripts, testing the commonsense reasoning ability of the models.
	\item LAMBADA~\citep{Gokaslan2019OpenWeb}: Evaluates the model's capability to predict the final word of a textual passage, focusing on the contextual understanding of the language.
\end{itemize}

For evaluation on downstream tasks, we adopted a few-shot approach, utilizing a 5-shot prompting method with greedy decoding and we report the average and variance of accuracy over 5 different seeds. Each prompt was structured with a task instruction followed by the five examples of training data and we computed accuracy on the validation set. We implemented our few-shot evaluation protocol following the guidelines provided by the lm\_evaluation\_harness library \footnote{\url{https://github.com/zphang/lm_evaluation_harness}}. For each configuration reported in the tables, the average GLUE score is computed first by taking the mean across all GLUE tasks. Subsequently, an overall average is calculated by averaging this GLUE score with the scores from ARC Easy, ARC Challenge, Hellaswag, and LAMBADA tasks.

\subsection{Comparison of Baseline and Pre-trained Model}

\begin{table}[!htb]
	\centering
	\caption{Comparison of baseline and pre-trained models: the latter trained for at least twice the duration.}
	\label{tab:base_line}
	\resizebox{\columnwidth}{!}{%
		\begin{tabular}{lcccc}
			\toprule
			            & WikiText103 & WikiText2 & PTB   & 1BW   \\ \midrule
			Baseline    & 39.94       & 34.32     & 35.13 & 44.03 \\
			Pre-trained & 29.17       & 24.67     & 35.86 & 45.87 \\ \bottomrule
		\end{tabular}%
	}
\end{table}

To establish a solid baseline for our experiments, we benchmark our trained model against the pre-trained GPT-2 weights provided by OpenAI across several downstream tasks. The comparison, detailed in Table~\ref{tab:base_line}, reveals that our model achieves results closely aligned with the original, validating the efficacy of our training approach. In the following sections, we will delve into the individual components of the model, discussing the impact and outcomes of our quantization experiments on each.

\section{Memory Analysis}
\label{app:memory_analysis}

\begin{figure*}[!htb]
	\centering
	\includegraphics[width=\linewidth]{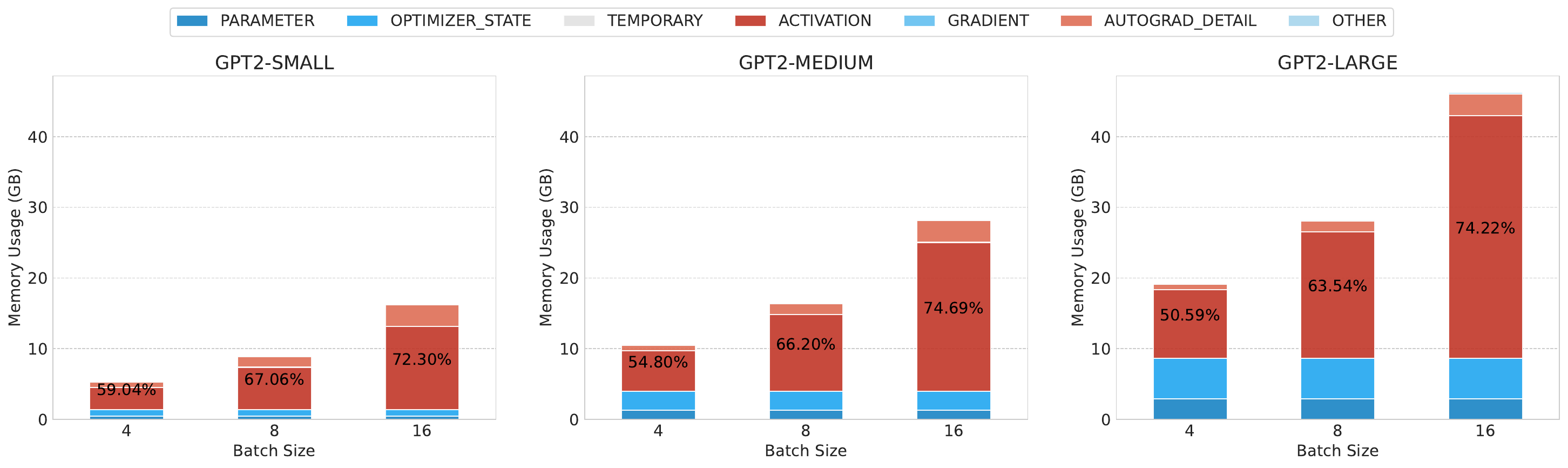}
	\caption{Distribution of peak memory usage across different model sizes (GPT-2 Small, Medium, and Large) for a constant context length of 1024, with varying batch sizes. This figure demonstrates how memory dedicated to activations increases as batch size increases, highlighting the impact of batch size on memory allocation dynamics in large-scale models.}
	\label{fig:memory_breakdown_batch}
\end{figure*}

\begin{figure*}[!htb]
	\centering
	\includegraphics[width=\linewidth]{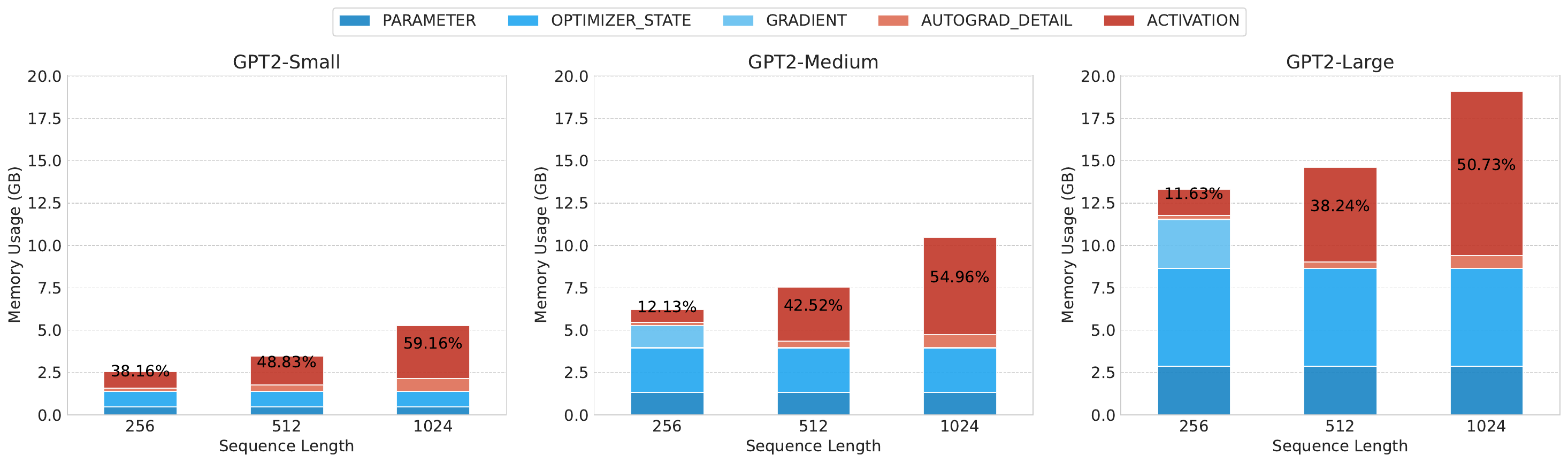}
	\caption{Peak memory usage profile for different sequence lengths while maintaining a constant batch size of 4, across different model sizes (GPT-2 Small, Medium, and Large). This figure illustrates the shift in peak memory usage from gradients to activations as sequence length increases and the significant impact of sequence length on memory dynamics during training.}
	\label{fig:memory_breakdown_seq}
\end{figure*}

In this section, we explore the memory consumption patterns of various components within GPT-2 models during training using the PyTorch Memory Profiler. This profiling tool allows for precise monitoring of memory usage throughout the lifecycle of a model's operation, particularly during the forward and backward passes and during optimization steps. "Peak memory" in this context refers to the maximum memory usage observed at any point in these stages, providing insights into how different model configurations impact overall memory requirements. We profiled all attention blocks as well as the LM head of the Transformer model.

Figure \ref{fig:memory_breakdown_batch} shows how memory usage changes with different batch sizes for a fixed context length of 1024 across various model sizes. Notably, as the batch size increases, the memory allocated for activations becomes more dominant, especially in larger models. This trend is primarily due to the need to store activations for the computation of gradients during the backward pass, which increases with larger batch sizes.

Figure \ref{fig:memory_breakdown_seq} examines the memory usage across various sequence lengths while keeping the batch size constant at 4. When both batch size and sequence length are small, peak memory typically occurs towards the end of the backward propagation phase. At this stage, memory includes the parameters, optimizer states, gradients from all layers, and activations from the initial layers. However, as the sequence length and batch size increase, peak memory usage shifts to the beginning of the backward propagation. At this point, the memory comprises parameters, optimizer states, all activations, and notably the output gradient of the final layers, which matches the size of the logits (proportional to batch size * sequence length * vocabulary size).

In conclusion, the analysis reveals that when a model can fit within the GPU memory, the majority of the memory at peak times is consumed by activations, especially when working with sufficiently large batch sizes and sequence lengths. Under these conditions, gradients do not contribute to peak memory usage. Consequently, quantizing gradients will not lead to significant memory savings.

\section{Quantization Results}
\label{app:quantization_results}

\begin{table*}[!htb]
	\centering
	\caption{\textbf{Weight Quantization}: Evaluation of perplexity across multiple datasets.}
	\label{tab:weight_perplexity_scores}
	\resizebox{0.5\linewidth}{!}{%
		\begin{tabular}{cc cccc}
			\toprule
			\#bit                  & granularity & WikiText103 & WikiText2 & PTB   & 1BW   \\
			\midrule
			\multicolumn{2}{c}{baseline} & 39.94 & 34.32 & 35.13 & 44.03 \\
			\midrule
			\multirow{2}{*}{4 bit} & per-tensor  & 55.50       & 46.70     & 52.38 & 59.14 \\
			                       & per-channel & 56.43       & 47.32     & 38.18 & 46.30 \\
			\midrule
			\multirow{2}{*}{8 bit} & per-tensor  & 48.52       & 40.01     & 37.02 & 45.04 \\
			                       & per-channel & 42.43       & 35.94     & 34.81 & 43.47 \\
			\bottomrule
		\end{tabular}
	}
\end{table*}

\begin{table*}[!htb]
	\centering
	\caption{\textbf{Activation Quantization}: Evaluation of perplexity across multiple datasets.}
	\label{tab:activation_perplexity_scores}
	\resizebox{0.5\linewidth}{!}{%
		\begin{tabular}{cc cccc}
			\toprule
			\#bit                  & granularity & WikiText103 & WikiText2 & PTB    & 1BW    \\
			\midrule
			\multicolumn{2}{c}{baseline} & 39.94 & 34.32 & 35.13 & 44.03 \\
			\midrule
			\multirow{2}{*}{4 bit} & per-tensor  & 418.63      & 264.07    & 261.64 & 310.43 \\
			                       & per-token   & 69.82       & 54.53     & 56.58  & 72.06  \\
			\midrule
			\multirow{2}{*}{8 bit} & per-tensor  & 64.77       & 51.06     & 37.96  & 45.26  \\
			                       & per-token   & 42.86       & 37.17     & 35.43  & 43.38  \\
			\bottomrule
		\end{tabular}
	}
\end{table*}

\begin{table*}[!htb]
	\centering
	\caption{\textbf{Gradient Quantization}: Evaluation of perplexity across multiple datasets.}
	\label{tab:gradient_perplexity_scores}
	\resizebox{0.5\linewidth}{!}{%
		\begin{tabular}{cc cccc}
			\toprule
			\#bit                  & granularity & WikiText103 & WikiText2 & PTB     & 1BW     \\
			\midrule
			\multicolumn{2}{c}{baseline} & 39.94 & 34.32 & 35.13 & 44.03 \\
			\midrule
			\multirow{2}{*}{4 bit} & per-tensor  & 17990.70    & 15560.03  & 6632.20 & 6393.07 \\
			                       & per-token   & 128.71      & 92.74     & 106.73  & 110.14  \\
			\midrule
			\multirow{2}{*}{8 bit} & per-tensor  & 123.08      & 87.81     & 104.90  & 111.51  \\
			                       & per-token   & 59.24       & 47.50     & 42.28   & 51.89   \\
			\bottomrule
		\end{tabular}
	}
\end{table*}

\begin{table*}[!htb]
	\centering
	\caption{\textbf{Adam Optimizer's First Moments}: Evaluation of perplexity across multiple datasets.}
	\label{tab:adam_first_order_moments}
	\resizebox{0.5\linewidth}{!}{%
		\begin{tabular}{cc cccc}
			\toprule
			\#bit                  & granularity & WikiText103 & WikiText2 & PTB   & 1BW   \\
			\midrule
			\multicolumn{2}{c}{baseline} & 39.94 & 34.32 & 35.13 & 44.03 \\
			\midrule
			\multirow{2}{*}{4 bit} & per-tensor  & 78.78       & 62.08     & 66.50 & 85.60 \\
			                       & per-channel & 43.02       & 36.70     & 38.57 & 47.90 \\
			\midrule
			\multirow{2}{*}{8 bit} & per-tensor  & 42.93       & 36.91     & 39.72 & 46.63 \\
			                       & per-channel & 39.84       & 33.78     & 35.67 & 44.29 \\
			\bottomrule
		\end{tabular}
	}
\end{table*}

\begin{table*}[!htb]
	\centering
	\caption{\textbf{Weight Quantization}: Few-shot accuracy on downstream tasks.}
	\label{tab:weight_all_scores}
	\resizebox{\linewidth}{!}{%
		\begin{tabular}{cc cccccc cc ccc c}
			\toprule
			\multicolumn{2}{c}{} & \multicolumn{6}{c}{GLUE Score} & \multicolumn{2}{c}{ARC} & \multicolumn{3}{c}{} & \multicolumn{1}{c} {} \\
			\cmidrule(lr){3-8} \cmidrule(lr){9-10} 
			\# of bit & granularity & MNLI & MRPC & RTE & QNLI & SST & WNLI & Easy & Challenge & Hellaswag & LAMBADA & \textbf{Average} &   \\
			\midrule
			\multicolumn{2}{c}{baseline} & $33.3_{\pm 0.4}$ & $61.7_{\pm 1.3}$ & $49.7_{\pm 2.6}$ & $49.2_{\pm 0.3}$ & $53.8_{\pm 1.9}$ & $46.8_{\pm 3.9}$ & $45.7_{\pm 0.3}$ & $22.5_{\pm 1.1}$ & $28.9_{\pm 0.1}$ & $36.17$ & $36.46$ \\
			\midrule
			\multirow{2}{*}{4 bit} & per-tensor & $32.1_{\pm 0.2}$ & $46.1_{\pm 1.2}$ & $49.0_{\pm 3.1}$ & $49.1_{\pm 0.2}$ & $54.1_{\pm 1.6}$ & $35.5_{\pm 4.2}$ & $39.9_{\pm 0.7}$ & $19.3_{\pm 0.5}$ & $27.2_{\pm 0.1}$ & $27.03$ & $31.54$ \\
			& per-channel & $33.3_{\pm 0.1}$ & $53.1_{\pm 2.2}$ & $49.7_{\pm 2.8}$ & $49.9_{\pm 0.5}$ & $54.6_{\pm 0.8}$ & $45.9_{\pm 4.6}$ & $44.6_{\pm 0.9}$ & $22.5_{\pm 0.7}$ & $28.7_{\pm 0.1}$ & $34.21$ & $35.55$ \\
			\midrule
			\multirow{2}{*}{8 bit} & per-tensor & $33.2_{\pm 0.3}$ & $59.1_{\pm 1.3}$ & $49.7_{\pm 2.6}$ & $49.1_{\pm 0.1}$ & $56.9_{\pm 2.6}$ & $41.1_{\pm 3.5}$ & $45.5_{\pm 0.8}$ & $20.4_{\pm 1.4}$ & $28.7_{\pm 0.2}$ & $34.81$ & $35.53$ \\
			& per-channel & $34.9_{\pm 0.3}$ & $62.7_{\pm 1.5}$ & $53.2_{\pm 1.3}$ & $49.5_{\pm 0.1}$ & $56.9_{\pm 1.1}$ & $53.5_{\pm 3.6}$ & $44.6_{\pm 0.5}$ & $21.1_{\pm 0.9}$ & $29.1_{\pm 0.2}$ & $36.43$ & $36.59$ \\
			\bottomrule
		\end{tabular}
	}
\end{table*}

\begin{table*}[!htb]
	\centering
	\caption{\textbf{Activation Quantization}: Few-shot accuracy on downstream tasks.}
	\label{tab:activation_all_scores}
	\resizebox{\linewidth}{!}{%
		\begin{tabular}{cc cccccc cc ccc c}
			\toprule
			\multicolumn{2}{c}{} & \multicolumn{6}{c}{GLUE Score} & \multicolumn{2}{c}{ARC} & \multicolumn{3}{c}{} & \multicolumn{1}{c} {} \\
			\cmidrule(lr){3-8} \cmidrule(lr){9-10} 
			\# of bit & granularity & MNLI & MRPC & RTE & QNLI & SST & WNLI & Easy & Challenge & Hellaswag & LAMBADA & \textbf{Average} &   \\
			\midrule
			\multicolumn{2}{c}{baseline} & $33.3_{\pm 0.4}$ & $61.7_{\pm 1.3}$ & $49.7_{\pm 2.6}$ & $49.2_{\pm 0.3}$ & $53.8_{\pm 1.9}$ & $46.8_{\pm 3.9}$ & $45.7_{\pm 0.3}$ & $22.5_{\pm 1.1}$ & $28.9_{\pm 0.1}$ & $36.17$ & $36.46$ \\
			\midrule
			\multirow{2}{*}{4 bit} & per-tensor & $34.6_{\pm 0.2}$ & $31.8_{\pm 0.2}$ & $50.5_{\pm 2.0}$ & $49.5_{\pm 0.0}$ & $50.2_{\pm 1.0}$ & $52.1_{\pm 2.7}$ & $30.4_{\pm 1.3}$ & $20.6_{\pm 2.6}$ & $26.1_{\pm 0.1}$ & $1.53$ & $24.67$ \\
			& per-token & $34.2_{\pm 0.4}$ & $58.3_{\pm 1.4}$ & $51.2_{\pm 2.3}$ & $49.3_{\pm 0.4}$ & $50.8_{\pm 0.3}$ & $52.4_{\pm 3.1}$ & $37.1_{\pm 1.6}$ & $21.0_{\pm 1.8}$ & $26.7_{\pm 0.1}$ & $18.88$ & $30.62$ \\
			& per-token asymmetric & $34.3_{\pm 0.2}$ & $54.1_{\pm 2.0}$ & $50.7_{\pm 2.9}$ & $49.1_{\pm 0.5}$ & $53.0_{\pm 1.4}$ & $55.2_{\pm 4.6}$ & $38.6_{\pm 0.9}$ & $18.9_{\pm 0.3}$ & $26.9_{\pm 0.2}$ & $22.30$ & $31.22$ \\
			\midrule
			\multirow{2}{*}{8 bit} & per-tensor & $32.7_{\pm 0.2}$ & $48.3_{\pm 1.4}$ & $47.1_{\pm 2.1}$ & $49.4_{\pm 0.3}$ & $54.6_{\pm 1.4}$ & $39.7_{\pm 3.0}$ & $44.1_{\pm 0.8}$ & $22.3_{\pm 1.5}$ & $28.8_{\pm 0.1}$ & $34.33$ & $34.97$ \\
			& per-token & $33.8_{\pm 0.4}$ & $56.2_{\pm 0.7}$ & $50.5_{\pm 3.4}$ & $49.8_{\pm 0.3}$ & $56.4_{\pm 1.5}$ & $54.1_{\pm 3.0}$ & $44.5_{\pm 0.6}$ & $20.0_{\pm 1.7}$ & $29.3_{\pm 0.1}$ & $36.46$ & $36.08$ \\
			\bottomrule
		\end{tabular}
	}
\end{table*}

\begin{table*}[!htb]
	\centering
	\caption{\textbf{Gradient Quantization}: Few-shot accuracy on downstream tasks.}
	\label{tab:grad_all_scores}
	\resizebox{\linewidth}{!}{%
		\begin{tabular}{cc cccccc cc ccc c}
			\toprule
			\multicolumn{2}{c}{} & \multicolumn{6}{c}{GLUE Score} & \multicolumn{2}{c}{ARC} & \multicolumn{3}{c}{} & \multicolumn{1}{c} {} \\
			\cmidrule(lr){3-8} \cmidrule(lr){9-10} 
			\# of bit & granularity & MNLI & MRPC & RTE & QNLI & SST & WNLI & Easy & Challenge & Hellaswag & LAMBADA & \textbf{Average} &   \\
			\midrule
			\multicolumn{2}{c}{baseline} & $33.3_{\pm 0.4}$ & $61.7_{\pm 1.3}$ & $49.7_{\pm 2.6}$ & $49.2_{\pm 0.3}$ & $53.8_{\pm 1.9}$ & $46.8_{\pm 3.9}$ & $45.7_{\pm 0.3}$ & $22.5_{\pm 1.1}$ & $28.9_{\pm 0.1}$ & $36.17$ & $36.46$ \\
			\midrule
			4 bit & per-token & $33.4_{\pm 0.4}$ & $36.6_{\pm 0.6}$ & $47.4_{\pm 2.6}$ & $49.5_{\pm 0.0}$ & $50.5_{\pm 0.8}$ & $58.3_{\pm 2.6}$ & $33.9_{\pm 1.0}$ & $20.2_{\pm 1.0}$ & $26.1_{\pm 0.1}$ & $19.06$ & $29.04$ \\
			\midrule
			\multirow{2}{*}{8 bit} & per-tensor & $34.2_{\pm 0.6}$ & $32.7_{\pm 0.6}$ & $50.3_{\pm 1.2}$ & $49.4_{\pm 0.0}$ & $50.5_{\pm 0.7}$ & $56.3_{\pm 2.0}$ & $32.1_{\pm 0.7}$ & $21.9_{\pm 1.0}$ & $26.3_{\pm 0.1}$ & $18.22$ & $28.81$ \\
			& per-token & $33.2_{\pm 0.1}$ & $42.6_{\pm 2.4}$ & $51.6_{\pm 1.8}$ & $49.5_{\pm 0.5}$ & $53.0_{\pm 2.0}$ & $22.3_{\pm 3.1}$ & $45.0_{\pm 1.0}$ & $19.5_{\pm 1.3}$ & $28.5_{\pm 0.1}$ & $33.81$ & $33.77$ \\
			\bottomrule
		\end{tabular}
	}
\end{table*}

\begin{table*}[!htb]
	\centering
	\caption{\textbf{Adam Optimizer’s First-Order Moments Quantization}: Few-shot accuracy on downstream tasks.}
	\label{tab:first_state_all_scores}
	\resizebox{\linewidth}{!}{%
		\begin{tabular}{cc cccccc cc ccc c}
			\toprule
			\multicolumn{2}{c}{} & \multicolumn{6}{c}{GLUE Score} & \multicolumn{2}{c}{ARC} & \multicolumn{3}{c}{} & \multicolumn{1}{c} {} \\
			\cmidrule(lr){3-8} \cmidrule(lr){9-10} 
			\# of bit & granularity & MNLI & MRPC & RTE & QNLI & SST & WNLI & Easy & Challenge & Hellaswag & LAMBADA & \textbf{Average} &   \\
			\midrule
			\multicolumn{2}{c}{baseline} & $33.3_{\pm 0.4}$ & $61.7_{\pm 1.3}$ & $49.7_{\pm 2.6}$ & $49.2_{\pm 0.3}$ & $53.8_{\pm 1.9}$ & $46.8_{\pm 3.9}$ & $45.7_{\pm 0.3}$ & $22.5_{\pm 1.1}$ & $28.9_{\pm 0.1}$ & $36.17$ & $36.46$ \\
			\midrule
			\multirow{2}{*}{4 bit} & per-tensor & $32.2_{\pm 0.3}$ & $64.9_{\pm 1.6}$ & $47.7_{\pm 0.6}$ & $50.2_{\pm 0.5}$ & $51.6_{\pm 1.7}$ & $38.9_{\pm 1.7}$ & $35.1_{\pm 1.6}$ & $20.1_{\pm 0.8}$ & $26.5_{\pm 0.2}$ & $20.12$ & $29.87$ \\
			& per-column & $33.8_{\pm 0.1}$ & $66.6_{\pm 1.2}$ & $50.0_{\pm 3.0}$ & $49.5_{\pm 0.0}$ & $53.5_{\pm 0.9}$ & $50.4_{\pm 2.7}$ & $44.6_{\pm 1.6}$ & $20.6_{\pm 0.9}$ & $28.3_{\pm 0.1}$ & $33.15$ & $35.44$ \\
			\midrule
			\multirow{2}{*}{8 bit} & per-tensor & $33.7_{\pm 0.4}$ & $59.2_{\pm 2.5}$ & $51.4_{\pm 2.7}$ & $50.1_{\pm 0.6}$ & $56.1_{\pm 1.4}$ & $51.8_{\pm 3.3}$ & $44.4_{\pm 0.7}$ & $18.5_{\pm 0.6}$ & $28.7_{\pm 0.1}$ & $32.91$ & $34.97$ \\
			& per-column & $33.3_{\pm 0.2}$ & $67.1_{\pm 1.0}$ & $50.8_{\pm 1.5}$ & $49.5_{\pm 0.0}$ & $55.3_{\pm 1.1}$ & $48.2_{\pm 2.1}$ & $43.8_{\pm 1.7}$ & $21.7_{\pm 1.3}$ & $28.9_{\pm 0.1}$ & $36.52$ & $36.34$ \\
			\bottomrule
		\end{tabular}
	}
\end{table*}

This appendix presents the granular results from our experiments on weight, activation, gradient, and optimizer states quantization. The tables detail the performance metrics, like perplexity and accuracy, under various quantization settings. Finally, we present post-training quantization results of our baseline model in Tables~\ref{tab:evaluation-ckps-ptq} and \ref{tab:evaluation-ckps-ptq-aq}.

\begin{table*}[!htb]
	\centering
	\caption{Post-training weight quantization results.}
	\label{tab:evaluation-ckps-ptq}
	\resizebox{0.6\linewidth}{!}{%
		\begin{tabular}{llcccc}
			\toprule
			\multirow{2}{*}{\#bit} & \multirow{2}{*}{granularity} & WikiText103 & WikiText2 & PTB      & 1BW      \\
			                       &                              & (ppl)       & (ppl)     & (ppl)    & (ppl)    \\ \midrule
			\multicolumn{2}{c}{baseline} & 39.94 & 34.32 & 35.13 & 44.03 \\ \midrule
			\multirow{2}{*}{4 bit} & per-tensor                   & 16196.10    & 17256.89  & 17471.35 & 13761.79 \\
			                       & per-column                   & 98.39       & 75.56     & 81.28    & 94.40    \\ \midrule
			\multirow{2}{*}{8 bit} & per-tensor                   & 46.45       & 39.23     & 41.18    & 52.15    \\
			                       & per-column                   & 40.15       & 34.45     & 35.23    & 44.11    \\ \bottomrule
		\end{tabular}%
	}
\end{table*}

\begin{table*}[!htb]
	\centering
	\caption{Post-training activation quantization results.}
	\label{tab:evaluation-ckps-ptq-aq}
	\resizebox{0.6\linewidth}{!}{%
		\begin{tabular}{llcccc}
			\toprule
			\multirow{2}{*}{\#bit} & \multirow{2}{*}{granularity} & WikiText103 & WikiText2 & PTB      & 1BW     \\
			                       &                              & (ppl)       & (ppl)     & (ppl)    & (ppl)   \\ \midrule
			\multicolumn{2}{c}{baseline} & 39.94 & 34.32 & 35.13 & 44.03 \\ \midrule
			\multirow{2}{*}{4 bit} & per-tensor                   & -           & -         & -        & -       \\
			                       & per-token                    & 14022.78    & 17933.29  & 13392.28 & 8763.06 \\ \midrule
			\multirow{2}{*}{8 bit} & per-tensor                   & 70.07       & 58.45     & 64.99    & 149.35  \\
			                       & per-token                    & 40.09       & 34.44     & 35.43    & 44.37   \\ \bottomrule
		\end{tabular}%
	}
\end{table*}

\end{document}